\documentclass[sigconf]{acmart}

\newcommand{\withvar}[2]{%
  #1\raisebox{-0.65ex}{\hspace{0.2em}\scriptsize #2}%
}
\usepackage{graphicx}
\usepackage{tabularx}
\usepackage{multirow}
\usepackage{booktabs}
\usepackage{colortbl}
\usepackage{algorithm}
\usepackage{algpseudocode}
\usepackage{placeins}
\makeatletter
\providecommand{\toclevel@algorithm}{1}
\makeatother
\setcopyright{none}
\acmYear{2027}
\copyrightyear{2027}
\acmDOI{}
\acmISBN{}
\acmConference[WWW '27]{The Web Conference 2027}{May 10--14, 2027}{Dublin, Ireland}
\title{RouteDRAG: Learning to Route and Control Agentic Execution for Multimodal Document RAG}

\begin{document}


\author{Zekai Chen*}
\affiliation{%
  \institution{Beijing Institute of Technology}
  \city{Beijing}
  \country{China}
}

\author{Hao Zhu*}
\affiliation{%
  \institution{Beijing Institute of Technology}
  \city{Beijing}
  \country{China}
}

\author{Yu Liu}
\affiliation{%
  \institution{Chinese Academy of Sciences}
  \country{China}
}

\author{Xunkai Li}
\affiliation{%
  \institution{Beijing Institute of Technology}
  \city{Beijing}
  \country{China}
}

\author{Zejian Kang}
\affiliation{%
  \institution{Zhejiang University}
  \city{Hangzhou}
  \country{China}
}
\affiliation{%
  \institution{Westlake University}
  \city{Hangzhou}
  \country{China}
}

\author{Rui Qian}
\affiliation{%
  \institution{Fudan University}
  \city{Shanghai}
  \country{China}
}

\author{Quanjiang Li}
\affiliation{%
  \institution{National University of Defense Technology}
  \city{Changsha}
  \country{China}
}

\author{Rong-Hua Li}
\affiliation{%
  \institution{Beijing Institute of Technology}
  \city{Beijing}
  \country{China}
}


\begin{abstract}
Web-based applications increasingly rely on understanding long multimodal documents, where answering questions often requires combining information across pages, tables, and figures. Agentic retrieval-augmented generation addresses these tasks through iterative retrieval and reasoning. However, Agentic execution is not always necessary, introducing additional computation and potentially incorrect answers. Our empirical analysis reveals that Agentic execution provides query-dependent benefits, while additional operations may improve, preserve, or degrade answer accuracy. These findings motivate controlling both Agent invocation and subsequent execution. To address these challenges, we propose \textbf{RouteDRAG}, a learning-based framework comprising four components. \textbf{Unified Document Index} organizes multimodal content into a graph-backed index and derives query representations from initial retrieval. \textbf{Adaptive Path Routing} preserves exact answering and predicts the relative benefit of Agent over shorter execution paths. \textbf{Dynamic Agent Execution} maintains an evolving workspace and refines Planner proposals to control tool use and termination. \textbf{Sequential Policy Learning} first trains operation control from observed outcomes and then trains path routing using the resulting Agent's performance. Across two benchmarks, RouteDRAG achieves an average accuracy improvement of 3.47 percentage points over shorter-path-only execution while assigning only 79.3\% of queries to Agent. Notably, it also surpasses uniform Agent execution on both benchmarks while requiring fewer answering calls. Our code is now available on \url{https://github.com/Ereshkigal1/RouteDRAG}.
\end{abstract}


\maketitle

\section{Introduction}
\label{sec:introduction}

Web-based applications, such as financial analysis platforms~\cite{chen2021finqa,zhu2021tatqa}, government information portals, academic search engines~\cite{dasigi2021qasper}, and online technical support~\cite{castelli2020techqa,feng2020doc2dial}, increasingly rely on understanding long and visually rich documents. Answering questions often requires locating and combining information distributed across pages, tables, and figures. For example, explaining a company's financial performance may require examining numerical results in a table and their explanations on other pages. A single retrieval step may locate only part of the required information, leaving the question unanswered. Agentic retrieval-augmented generation (Agentic RAG) addresses this challenge through iterative retrieval, visual understanding, and multi-step reasoning, allowing information from different document regions to be located and combined.

\begin{figure*}[t]
    \centering
    \includegraphics[width=\textwidth]{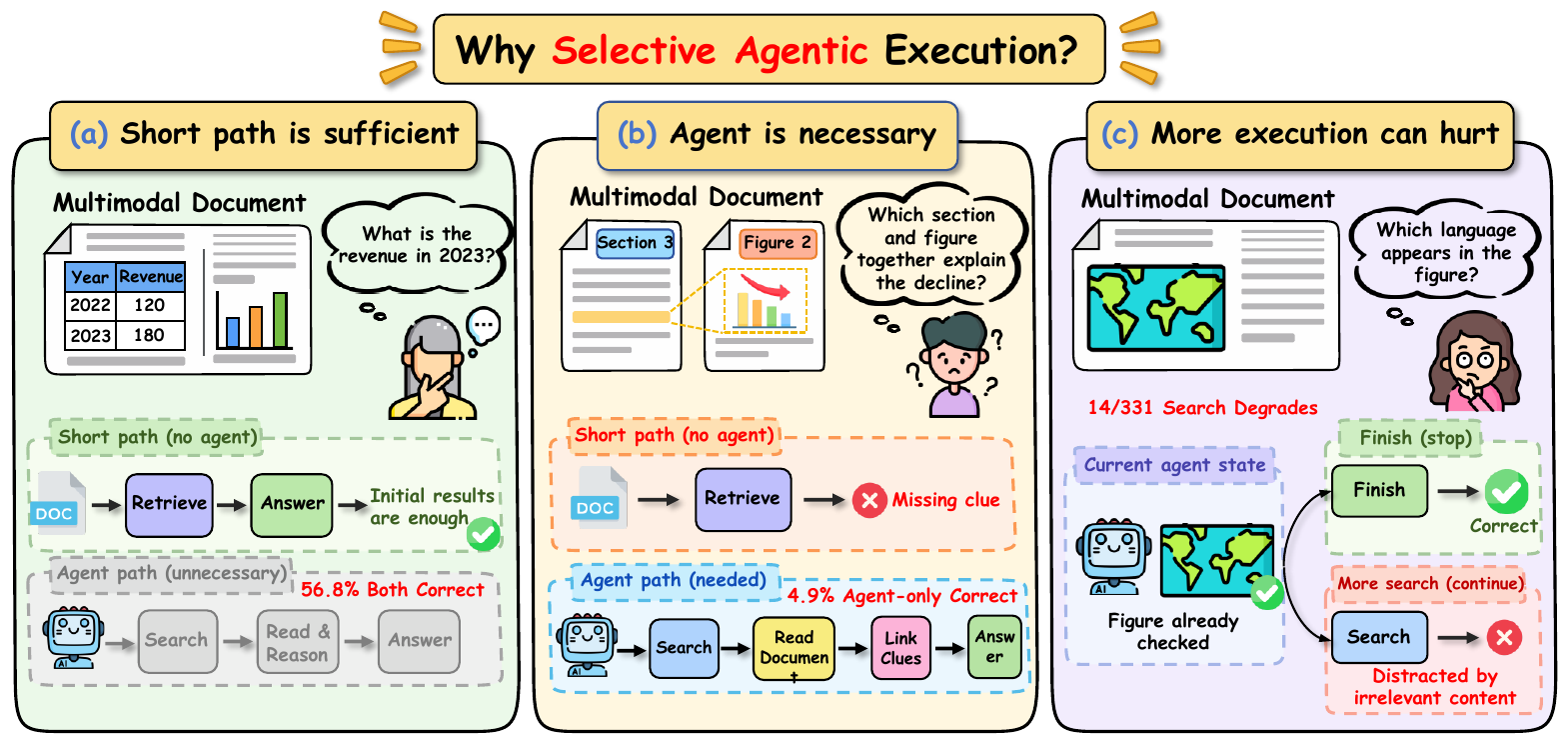}
    \caption{Why selective Agentic execution? (a) Initial retrieval can suffice, making Agent execution redundant. (b) Others require linking information through Agent execution. (c) Further retrieval can distract an Agent with sufficient information.}
    \label{fig:motivation}
\end{figure*}

Recent studies have made Agentic RAG more adaptive through multimodal retrieval, graph-based exploration and iterative reasoning~\cite{wang2025vidorag,sourati2026ladrag,zuo2026magerag,ji2026taprag,liu2026opera}. These methods allow Agents to dynamically select document content and adjust retrieval procedures. However, adapting how information is retrieved does not fully address how much Agentic execution is needed. As illustrated in Fig.~\ref{fig:motivation}, some questions require multiple retrieval steps to locate and combine information across different pages, while others can be answered directly from the initial retrieval results without invoking an Agent. Unnecessary Agent invocation increases computational cost and may introduce irrelevant or conflicting information, leading to incorrect answers. Yet deciding whether to invoke an Agent is only the first step. As the Agent gathers more information, it must further decide whether to retrieve additional content, inspect visual regions, or terminate execution. This raises two fundamental questions: \emph{when should Agentic execution be invoked, and how should its subsequent operations be controlled?}

Our empirical analysis reveals three observations about Agentic execution.
\textbf{O1: Limited Overall Gains.}
Agent improves accuracy by 1.00 and 3.12 percentage points on DocBench and MMLongBench-Doc, respectively, but uses $1.65\times$ and $1.81\times$ as many answering tokens as Standard.
\textbf{O2: Query-Dependent Benefits.}
Agent alone answers 3.09\% and 6.69\% of questions correctly, while both Agent and Standard succeed on 69.24\% and 44.18\%. Conversely, Standard alone succeeds on 2.09\% and 3.57\%, indicating that Agent execution is not universally beneficial.
\textbf{O3: Non-Monotonic Operation Gains.}
Among 917 matched operation-level comparisons, 84.51\% show unchanged correctness, while additional operations can improve or degrade answers.
These observations motivate selective Agent invocation based on its expected benefit and adaptive operation control based on information collected during execution.

To address these challenges, we propose \textbf{RouteDRAG}, a learning-based framework for selective Agentic document RAG. First, \textbf{Unified Document Index} organizes multimodal content into a graph-backed index and derives query representations from initial retrieval. Next, \textbf{Adaptive Path Routing} predicts the benefit of invoking Agent over a shorter path while preserving exact answering. When Agent is invoked, \textbf{Dynamic Agent Execution} maintains an evolving workspace and refines Planner proposals to control subsequent operations and termination. Finally, \textbf{Sequential Policy Learning} first trains operation control from observed outcomes and then trains path routing using the resulting Agent's performance. Together, these components adapt Agent invocation and execution to individual queries.

\textbf{Our Contributions.}
\underline{\textit{(1) New Perspective.}} We identify limited overall gains, query-dependent benefits, and non-monotonic operation gains in Agentic document RAG, motivating execution decisions at both query and operation levels.
\underline{\textit{(2) New Framework.}} We propose \textbf{RouteDRAG}, integrating Unified Document Index, Adaptive Path Routing, Dynamic Agent Execution, and Sequential Policy Learning to control Agent invocation and subsequent operations.
\underline{\textit{(3) SOTA Performance.}} We evaluate RouteDRAG against nine RAG methods on DocBench and MMLongBench-Doc, achieving 75.86\% and 53.80\% accuracy, respectively, while reducing answering calls by 16.4\% and 14.9\% compared with uniform Agent execution.

\section{Related Work}
\label{sec:related_work}

\subsection{Multimodal and Agentic Document RAG}

Document RAG has evolved from text-based retrieval toward richer multimodal and structural representations. LightRAG~\cite{guo2025lightrag} organizes entities and relations into a graph, while RAG-Anything~\cite{guo2025raganything} extends graph-based retrieval to multimodal documents.
GraphReader \cite{li2024graphreader} further employs graph-guided Agent exploration for long-text understanding.

G$^2$-Reader \cite{du2026g2reader} maintains evolving content and planning graphs for multimodal document question answering.

 VisRAG~\cite{yu2025visrag} and M3DocRAG~\cite{cho2024m3docrag} further explore visual retrieval for document understanding. 
 
AgenticOCR \cite{wang2026agenticocr} enables query-driven parsing of relevant visual regions to reduce redundant document processing.

ReAct \cite{yao2023react} established a reasoning-and-acting paradigm that interleaves language-model reasoning with external actions.

 Recent Agentic approaches support more flexible information acquisition. ViDoRAG~\cite{wang2025vidorag} employs iterative exploration and reasoning, LAD-RAG~\cite{sourati2026ladrag} combines neural and structural document retrieval, and MAGE-RAG~\cite{zuo2026magerag} dynamically explores and prunes graph-based evidence. TAP-RAG~\cite{ji2026taprag} further adapts graph expansion and visual inspection to query requirements. These methods improve the acquisition and organization of multimodal information. RouteDRAG instead focuses on whether such Agentic execution is necessary for a given query and how its subsequent operations should be controlled.

\subsection{Adaptive Retrieval and Execution Selection}

Adaptive RAG methods adjust retrieval and execution according to query requirements. Self-RAG~\cite{asai2024selfrag} learns when retrieval is needed, while Adaptive-RAG~\cite{jeong2024adaptiverag} selects among no-retrieval, single-step, and iterative strategies based on query complexity. 

Active Retrieval Augmented Generation \cite{jiang2023active} dynamically retrieves additional evidence during generation.

MBA-RAG~\cite{tang2025mbarag} further uses bandit-based selection to balance retrieval quality and efficiency. More recent methods extend adaptation to multi-step execution. MAO-ARAG~\cite{chen2025maoarag} dynamically organizes retrieval agents into query-specific workflows, while SPARKLE~\cite{fang2026sparkle} learns a separate policy for controlling retrieval operations. These studies demonstrate the value of both query-level adaptation and step-wise control. RouteDRAG connects the two decisions by estimating Agent's benefit relative to a designated shorter path and refining subsequent operations based on the evolving workspace. Its sequential learning process further aligns path routing with the behavior of the resulting Agent.
\begin{figure*}[tbp]
    \centering
    \includegraphics[width=\textwidth]{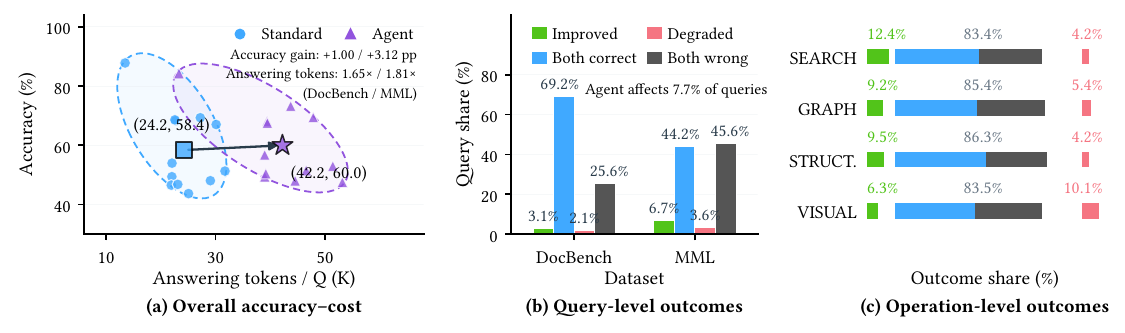}
    \caption{Empirical analysis of Agentic execution. (a) Accuracy and answering-token usage across document categories for Standard and Agent. (b) Paired query outcomes, illustrating the query-dependent benefits of Agentic execution. (c) Answer correctness changes after an additional operation compared with immediate termination.}
    \label{fig:empirical}
\end{figure*}
\section{Problem Definition}
\label{sec:problem}

Given a multimodal document $D$ and a query $q$, let $\mathcal{K}_D$ denote the unified document index. We consider four execution paths:
\begin{equation}
\mathcal{C} =
\{\text{Exact}, \text{Standard}, \text{Direct VLM}, \text{Agent}\},
\end{equation}
where the first three provide shorter answering procedures, while Agent supports iterative retrieval and reasoning.
Unlike execution-path selection alone, selective Agentic execution involves two related decisions. At the query level, a routing policy $\pi(q,\mathcal{K}_D)$ determines whether Agent is necessary or a shorter path is sufficient. Once Agent is invoked, its execution policy selects an operation $a_t$ from the legal action set $\Gamma_t$ based on the evolving workspace $W_t$, including whether to continue acquiring information or terminate. Our objective is to maintain answer accuracy while avoiding unnecessary Agent invocation and subsequent operations.

\section{Empirical Study}
\label{sec:empirical}

To understand when Agentic execution is necessary, we investigate its benefits from three perspectives. We use a standalone reference Agent for these analyses, distinct from the final RouteDRAG Actor evaluated in the main experiments. We first examine whether uniform Agent execution justifies its additional cost, then investigate how its benefits vary across queries, and finally assess whether additional operations consistently improve answers.

\textbf{O1: Limited Overall Gains.}
Does Agentic execution justify its additional computation? Fig.~\ref{fig:empirical}(a) compares Standard and Agent across five DocBench domains and seven MMLongBench-Doc document types. Agent improves accuracy by 1.00 and 3.12 percentage points on DocBench and MMLongBench-Doc, respectively, while consuming $1.65\times$ and $1.81\times$ the answering tokens. These results reveal a mismatch between overall accuracy gains and computational cost: uniformly invoking the Agent incurs substantial overhead for modest improvements. However, aggregate results cannot identify which queries justify this additional computation, motivating a closer examination of query-level benefits.

\textbf{O2: Query-Dependent Benefits.}
Which queries actually benefit from Agent execution? Fig.~\ref{fig:empirical}(b) compares the two paths on the same questions. Correctness remains unchanged for 94.83\% of DocBench queries and 89.73\% of MMLongBench-Doc queries. Meanwhile, Agent recovers answers missed by Standard on 3.09\% and 6.69\% of queries, but also turns correct answers into incorrect ones on 2.09\% and 3.57\%, respectively. Thus, Agent execution is neither universally necessary nor uniformly beneficial. Its value depends on the query, motivating \emph{Adaptive Path Routing} to estimate the expected benefit of Agent execution over the designated short path.

\textbf{O3: Non-Monotonic Operation Gains.}
Even when Agent execution is warranted, does acquiring more information necessarily improve the answer? We examine 917 matched comparisons between immediate FINISH and executing one additional operation before FINISH. As shown in Fig.~\ref{fig:empirical}(c), 84.51\% of comparisons leave correctness unchanged, while the remaining operations can either improve or degrade answers. In particular, VISUAL operations degrade correctness more often than they improve it, suggesting that additional information may introduce noise rather than useful evidence. This motivates \emph{Dynamic Agent Execution}, which adapts operation selection to the evolving workspace instead of assuming that continued execution is always beneficial.

\begin{figure*}[t]
\centering
\includegraphics[width=\textwidth,trim=40 120 245 105,clip]{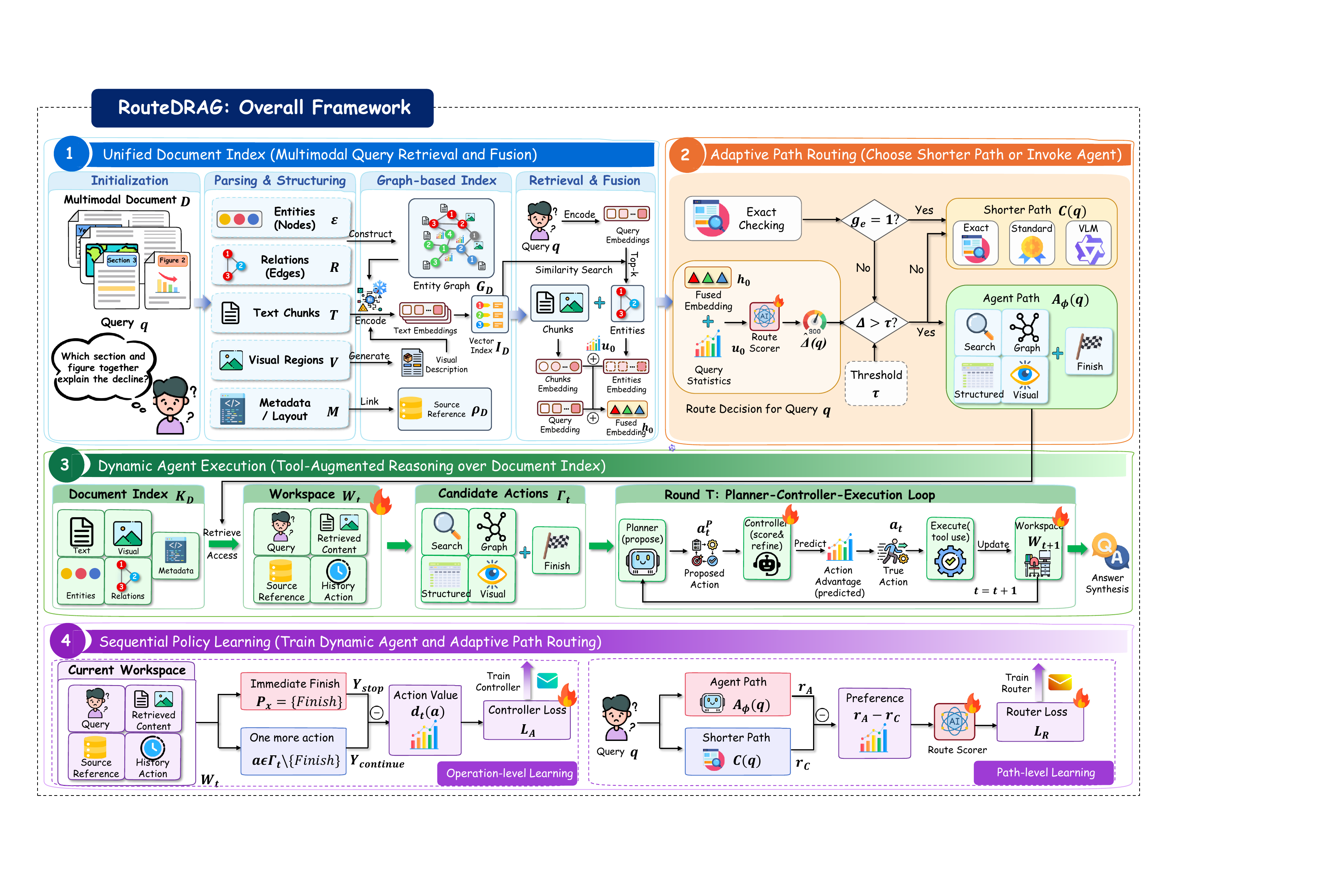}
\caption{Overview of RouteDRAG. A shared document index supports learned path routing and Agent execution control. Actor-aligned policy learning uses paired execution outcomes to train routing and action decisions in stages.}
\label{fig:routedrag_framework}
\end{figure*}

\section{Methodology}
\label{sec:method}

\subsection{Overview}
\label{sec:method_overview}

Fig.~\ref{fig:routedrag_framework} illustrates the overall framework. RouteDRAG addresses selective Agentic execution through two related decisions: when to invoke Agent and how to control its subsequent operations. These decisions are learned from observed execution outcomes at the query and operation levels. The framework comprises four components: (1) Unified Document Index, (2) Adaptive Path Routing, (3) Dynamic Agent Execution, and (4) Sequential Policy Learning.

\subsection{Unified Document Index}
\label{sec:shared_index}

\textbf{Motivation.} Effective path routing requires considering not only the query but also the document content retrieved before execution. We therefore construct a shared document index and learn a query-conditioned representation that can be reused during execution.

\textbf{Graph-Backed Document Organization.} Following RAG-Anything and LightRAG~\cite{guo2025raganything,guo2025lightrag}, we organize each multimodal document into the shared index $\mathcal{K}_D$ defined in the problem formulation. The index connects text chunks, entities, relations, visual elements, and structured metadata with their original document locations. It supports four execution paths: Exact, Standard, Direct VLM, and Agent. The document index and pretrained retrieval encoder remain frozen during training.

\textbf{Query-Conditioned Representation.} Given a query $q$, we retrieve the most relevant text chunks and graph entities, retaining up to eight chunks and four entities. A frozen semantic encoder produces their embeddings, denoted by $\mathbf{e}_q$, $\mathbf{E}_c$, and $\mathbf{E}_g$. A shared trainable projection $P_{\psi}$ maps them into a 64-dimensional space. We then obtain query-conditioned summaries through masked attention:
\begin{equation}
\begin{aligned}
\mathbf{z}_q &= P_{\psi}(\mathbf{e}_q),
\mathbf{z}_c = \operatorname{Attn}_{\psi}
(\mathbf{z}_q,P_{\psi}(\mathbf{E}_c)),
\mathbf{z}_g = \operatorname{Attn}_{\psi}
(\mathbf{z}_q,P_{\psi}(\mathbf{E}_g)).
\end{aligned}
\label{eq:document_attention}
\end{equation}
Here, the attention operation excludes unavailable entries through a validity mask. To adapt the contribution of each source to the current query, we compute
\begin{equation}
\begin{aligned}
\boldsymbol{\alpha}
&=\operatorname{softmax}
\left(G_{\psi}([\mathbf{z}_q;\mathbf{z}_c;\mathbf{z}_g])\right),
\mathbf{h}_0
=F_{\psi}\left(
[\alpha_q\mathbf{z}_q;
\alpha_c\mathbf{z}_c;
\alpha_g\mathbf{z}_g;
U_{\psi}(\mathbf{u}_0)]
\right),
\end{aligned}
\label{eq:query_fusion}
\end{equation}
where $\mathbf{u}_0$ contains query characteristics, retrieved-content statistics, and similarity scores. $G_{\psi}$ and $F_{\psi}$ denote lightweight gating and fusion networks. The resulting representation $\mathbf{h}_0$ summarizes the query and initial retrieval results before Agent invocation.

\subsection{Adaptive Path Routing}
\label{sec:path_routing}

\textbf{Motivation.} A complex question does not necessarily require Agentic execution, while a seemingly simple question may need multiple retrieval operations. We therefore learn whether Agent improves upon an available non-Agent path, rather than predicting execution requirements solely from query types.

\textbf{Relative Path Prediction.} RouteDRAG first applies a verified Exact gate. Otherwise, a fixed policy selects the available shorter path $C(q)\in\{\mathrm{Standard},\mathrm{VLM}\}$. Instead of independently estimating every path's correctness, we predict the relative advantage of the Agent:
\begin{equation}
\widehat{\Delta}_{\theta}(q)
=
\mathbf{w}_h^{\top}\mathbf{h}_0
+
\mathbf{w}_u^{\top}\widetilde{\mathbf{u}}_0+b,
\label{eq:path_margin}
\end{equation}
where $\widetilde{\mathbf{u}}_0$ contains normalized query and retrieval statistics. The first term models interactions between the query and retrieved document content, while the second preserves direct access to lightweight descriptive features.

\textbf{Execution-Path Selection.} The resulting routing policy is
\begin{equation}
\pi_{\theta}(q)=
\begin{cases}
\mathrm{Exact},
& g_{\mathrm{exact}}(q,\mathcal{K}_D)=1,\\
\mathrm{Agent},
& \widehat{\Delta}_{\theta}(q)>\tau,\\
C(q),
& \text{otherwise},
\end{cases}
\label{eq:learned_routing}
\end{equation}
where $g_{\mathrm{exact}}$ verifies complete structured metadata and $\tau$ is determined on the development set. This formulation turns Agent invocation into a relative execution decision: the Agent is selected when its predicted improvement over the designated shorter path exceeds the routing threshold.

\subsection{Dynamic Agent Execution}
\label{sec:agent_control}

\textbf{Motivation.} Agent operations should adapt to the information accumulated during execution. We integrate learned action refinement with the LLM Planner, using the evolving execution state to guide its proposals.

\textbf{State-Conditioned Operation Modeling.} Agent starts with native hybrid retrieval and maintains a workspace $W_t$ containing retrieved content, source references, and operation history. The shared projection network produces its state representation $\mathbf{h}_t$. For each candidate operation $a$ with embedding $\mathbf{e}_a$, we compute action-conditioned summaries of text chunks and graph entities:
\begin{equation}
\mathbf{z}_{t,a}^{m}
=
\operatorname{Attn}_{\phi}
\left(
\mathbf{z}_q+\mathbf{e}_a,
P_{\psi}(\mathbf{E}_{t}^{m})
\right),
\quad m\in\{c,g\}.
\label{eq:action_conditioned_attention}
\end{equation}
Here, $\mathbf{E}_{t}^{c}$ and $\mathbf{E}_{t}^{g}$ denote current chunk and entity embeddings. We combine these summaries with the execution state and operation history:
\begin{equation}
z_{\phi}(W_t,a)
=
H_{\phi}\left(
[\mathbf{h}_t;
\mathbf{z}_{t,a}^{c};
\mathbf{z}_{t,a}^{g};
\mathbf{e}_a;
\boldsymbol{\ell}_t]
\right),
\label{eq:action_score}
\end{equation}
where $\boldsymbol{\ell}_t$ encodes previous operations. The relative operation score is defined against immediate termination:
\begin{equation}
\widehat{d}_t(a)
=
z_{\phi}(W_t,a)
-
z_{\phi}(W_t,\mathrm{FINISH}).
\label{eq:relative_action_score}
\end{equation}
This provides a common reference for comparing continued execution with termination.

\textbf{Constrained Action Refinement.} At each step, native execution rules determine the legal action set $\Gamma_t$, including retrieval, graph expansion, structured lookup, visual inspection, and \texttt{FINISH} when permitted. Given the Planner's proposal $a_t^P$, the controller identifies
\begin{equation}
a_t^*
=
\arg\max_{a\in\Gamma_t}
\widehat{d}_t(a).
\label{eq:preferred_action}
\end{equation}
The selected action is
\begin{equation}
a_t=
\begin{cases}
a_t^*,
& \widehat{d}_t(a_t^*)-
\widehat{d}_t(a_t^P)>m,\\
a_t^P,
& \text{otherwise},
\end{cases}
\label{eq:action_refinement}
\end{equation}
where $m$ is a fixed margin. This refinement applies to valid Planner proposals, while mandatory actions follow native rules. If the controller changes the operation, the Planner generates its arguments, preserving the original tool execution procedure.

\textbf{State Transition.} The selected operation returns $o_t$, updating the workspace as $W_{t+1}=\operatorname{Commit}(W_t,o_t)$. The update incorporates new content, preserves source references, and removes duplicates. Subsequent decisions use the updated workspace until \texttt{FINISH} or the execution limit is reached, followed by answer synthesis and validation.

\subsection{Sequential Policy Learning}
\label{sec:policy_learning}

\textbf{Motivation.} Agent invocation depends on its actual performance, while operation utility varies with the execution state. We therefore learn both decisions from paired outcomes, training the router after updating the Agent controller.

\textbf{Operation-Level Learning.} We compare immediate termination with one additional operation at observed Agent states. For state $W_{i,t}$ and operation $a$, the target is
\begin{equation}
d_{i,t,a}
=
Y\left(
\operatorname{Stop}(\mathcal{T}_a(W_{i,t}))
\right)
-
Y\left(
\operatorname{Stop}(W_{i,t})
\right),
\label{eq:operation_gain}
\end{equation}
where $Y(\cdot)\in\{0,1\}$ denotes correctness and $\operatorname{Stop}$ applies a fixed answer-generation procedure. Only evaluated state--action pairs are used.
The controller combines regression, action ranking, and Planner imitation:
\begin{equation}
\mathcal{L}_{A}
=
\mathcal{L}_{\mathrm{reg}}
+
\alpha\mathcal{L}_{\mathrm{rank}}^{A}
+
\beta\mathcal{L}_{\mathrm{BC}},
\label{eq:controller_loss}
\end{equation}
where $\mathcal{L}_{\mathrm{reg}}$ fits $\widehat{d}_t(a)$ to the observed gain, and $\mathcal{L}_{\mathrm{BC}}$ preserves Planner preferences. The ranking loss is
\begin{equation}
\mathcal{L}_{\mathrm{rank}}^{A}
=
\frac{1}{|\mathcal{P}_A|}
\sum_{(t,a,b)\in\mathcal{P}_A}
\operatorname{softplus}
\left(
-\left[\widehat{d}_t(a)-\widehat{d}_t(b)\right]
\right),
\label{eq:action_ranking}
\end{equation}
where $\mathcal{P}_A$ contains evaluated same-state action pairs with $d_t(a)>d_t(b)$.

\textbf{Path-Level Learning.} We execute the updated Agent $A_{\phi}$ and compare its correctness with the fixed shorter path:
\begin{equation}
\Delta_i
=
Y(A_{\phi}(q_i))
-
Y(C(q_i))
\in\{-1,0,1\}.
\label{eq:paired_route_target}
\end{equation}
The router predicts this relative gain using regression and cross-query ranking:
\begin{equation}
\mathcal{L}_{R}
=\mathbb{E}_{i}(\hat{\Delta}_i-\Delta_i)^2
+\mu\mathbb{E}_{(i,j)\in\mathcal{P}_R}
\ell(\hat{\Delta}_j-\hat{\Delta}_i).
\label{eq:router_loss}
\end{equation}
where $\hat{\Delta}_i=\widehat{\Delta}_{\theta}(q_i)$, $\ell(x)=\operatorname{softplus}(x)$, and $\mathcal{P}_R$ contains query pairs satisfying $\Delta_i>\Delta_j$.

\textbf{Sequential Training.} RouteDRAG first trains the shared representation and Agent controller from observed operation outcomes. It then freezes both and trains the routing head using the updated Agent's execution results. This ensures that path routing reflects the Agent's actual behavior. All parameters remain fixed during evaluation.

\begin{table*}[t]
\centering
\caption{Answer accuracy (\%) on MMLongBench-Doc. Best and second-best results are bold and underlined, respectively.}
\label{tab:canonical_mml}
\small
\setlength{\tabcolsep}{1.5pt}
\renewcommand{\arraystretch}{1.14}
\begin{tabularx}{\textwidth}{@{}>{\raggedright\arraybackslash}p{0.115\textwidth}>{\raggedright\arraybackslash}p{0.18\textwidth}*{9}{>{\centering\arraybackslash}X}@{}}
\toprule
& & \multicolumn{5}{c}{\textbf{Support source}} & \multicolumn{3}{c}{\textbf{Support requirement}} & \\
\cmidrule(lr){3-7}\cmidrule(lr){8-10}
\textbf{Type} & \textbf{Method} & \textbf{TXT} & \textbf{LAY} & \textbf{CHA} & \textbf{TAB} & \textbf{FIG} & \textbf{SIN} & \textbf{MUL} & \textbf{UNA} & \textbf{Overall} \\
\midrule
\shortstack[l]{Direct} & Direct MLLM
& \withvar{30.76}{1.88} & \withvar{36.30}{2.09} & \withvar{28.58}{1.72} & \withvar{20.52}{2.16} & \withvar{28.46}{1.94}
& \withvar{36.30}{1.66} & \withvar{21.36}{2.31} & \withvar{52.46}{1.83} & 35.00 \\
\midrule
\multirow{2}{*}{\shortstack[l]{Text RAG}} & BM25
& \withvar{34.86}{1.19} & \withvar{19.20}{1.37} & \withvar{17.33}{1.12} & \withvar{31.88}{1.48} & \withvar{14.22}{1.06}
& \withvar{34.00}{0.94} & \withvar{16.98}{1.56} & \withvar{\textbf{76.64}}{1.21} & 37.98 \\
& BGE-M3
& \withvar{33.72}{1.32} & \withvar{23.17}{1.18} & \withvar{13.96}{1.09} & \withvar{35.82}{1.41} & \withvar{16.15}{1.02}
& \withvar{33.15}{0.96} & \withvar{19.99}{1.53} & \withvar{\underline{73.36}}{1.24} & 37.86 \\
\midrule
\multirow{2}{*}{\shortstack[l]{Multimodal}} & M3DocRAG~\cite{cho2024m3docrag}
& \withvar{\underline{43.47}}{1.79} & \withvar{31.81}{1.63} & \withvar{\underline{39.23}}{1.51} & \withvar{33.45}{1.91} & \withvar{37.33}{1.43}
& \withvar{\underline{53.08}}{1.32} & \withvar{25.38}{1.72} & \withvar{25.00}{1.98} & 37.58 \\
& EVisRAG~\cite{sun2025visrag2}
& \withvar{38.29}{1.22} & \withvar{33.21}{1.34} & \withvar{38.17}{1.11} & \withvar{39.84}{1.46} & \withvar{\underline{38.90}}{0.98}
& \withvar{51.55}{0.92} & \withvar{25.03}{1.28} & \withvar{58.20}{1.57} & 44.27 \\
\midrule
\multirow{5}{*}{\shortstack[l]{Graph/Agentic}} & RAG-Anything~\cite{guo2025raganything}
& \withvar{40.00}{2.08} & \withvar{29.30}{1.91} & \withvar{29.21}{1.76} & \withvar{\underline{43.45}}{2.22} & \withvar{33.73}{1.59}
& \withvar{47.59}{1.53} & \withvar{\underline{26.93}}{1.82} & \withvar{60.25}{2.31} & 43.61 \\
& MoLoRAG~\cite{wu2025molorag}
& \withvar{38.88}{2.39} & \withvar{\underline{36.82}}{2.17} & \withvar{35.86}{1.94} & \withvar{37.55}{2.48} & \withvar{32.80}{2.08}
& \withvar{51.28}{1.85} & \withvar{22.31}{2.23} & \withvar{52.46}{2.61} & 41.96 \\
& ViDoRAG~\cite{wang2025vidorag}
& \withvar{38.21}{1.93} & \withvar{27.41}{1.78} & \withvar{27.45}{1.64} & \withvar{34.66}{2.02} & \withvar{33.37}{1.48}
& \withvar{48.69}{1.55} & \withvar{18.74}{1.86} & \withvar{62.70}{2.19} & 41.85 \\
& TAP-RAG~\cite{ji2026taprag}
& \withvar{43.25}{1.69} & \withvar{28.36}{1.55} & \withvar{34.91}{1.34} & \withvar{36.79}{1.82} & \withvar{36.22}{1.46}
& \withvar{52.20}{1.27} & \withvar{24.84}{1.61} & \withvar{60.45}{1.98} & \underline{45.01} \\
& \cellcolor{black!6}\textbf{RouteDRAG}
& \cellcolor{black!6}\withvar{\textbf{51.68}}{0.97}
& \cellcolor{black!6}\withvar{\textbf{46.22}}{1.08}
& \cellcolor{black!6}\withvar{\textbf{41.57}}{0.82}
& \cellcolor{black!6}\withvar{\textbf{55.05}}{1.19}
& \cellcolor{black!6}\withvar{\textbf{42.66}}{0.72}
& \cellcolor{black!6}\withvar{\textbf{58.75}}{0.61}
& \cellcolor{black!6}\withvar{\textbf{37.99}}{0.91}
& \cellcolor{black!6}\withvar{67.62}{1.31}
& \cellcolor{black!6}\textbf{53.80} \\
\bottomrule
\end{tabularx}
\end{table*}

\begin{table*}[t]
\centering
\caption{Answer accuracy (\%) on DocBench. Bold and underlined values indicate the best and second-best results, respectively.}
\label{tab:canonical_overall}
\small
\setlength{\tabcolsep}{1.5pt}
\renewcommand{\arraystretch}{1.14}
\begin{tabularx}{\textwidth}{@{}>{\raggedright\arraybackslash}p{0.115\textwidth}>{\raggedright\arraybackslash}p{0.18\textwidth}*{10}{>{\centering\arraybackslash}X}@{}}
\toprule
& & \multicolumn{5}{c}{\textbf{Document domain}} & \multicolumn{4}{c}{\textbf{Question type}} & \\
\cmidrule(lr){3-7}\cmidrule(lr){8-11}
\textbf{Type} & \textbf{Method} & \textbf{Acad.} & \textbf{Finance} & \textbf{Gov.} & \textbf{Law} & \textbf{News} & \textbf{TXT} & \textbf{MM} & \textbf{META} & \textbf{UNA} & \textbf{Overall} \\
\midrule
\shortstack[l]{Direct} & Direct MLLM
& \withvar{30.36}{2.14} & \withvar{39.58}{2.38} & \withvar{68.92}{1.92} & \withvar{65.97}{2.26} & \withvar{33.14}{2.47}
& \withvar{52.43}{1.85} & \withvar{22.40}{2.55} & \withvar{49.22}{2.08} & \withvar{63.71}{2.31} & 44.55 \\
\midrule
\multirow{2}{*}{\shortstack[l]{Text RAG}} & BM25
& \withvar{61.06}{1.45} & \withvar{57.64}{1.62} & \withvar{\underline{74.32}}{1.27} & \withvar{\underline{73.82}}{1.73} & \withvar{87.21}{1.18}
& \withvar{\underline{89.81}}{1.06} & \withvar{58.44}{1.84} & \withvar{39.53}{1.91} & \withvar{\underline{80.65}}{1.39} & 68.24 \\
& BGE-M3
& \withvar{55.78}{1.58} & \withvar{54.51}{1.42} & \withvar{67.57}{1.31} & \withvar{71.20}{1.64} & \withvar{86.63}{1.22}
& \withvar{82.77}{1.15} & \withvar{52.92}{1.78} & \withvar{42.25}{1.96} & \withvar{79.03}{1.37} & 64.52 \\
\midrule
\multirow{2}{*}{\shortstack[l]{Multimodal}} & M3DocRAG~\cite{cho2024m3docrag}
& \withvar{61.06}{2.03} & \withvar{65.28}{1.87} & \withvar{66.89}{1.74} & \withvar{71.20}{2.18} & \withvar{81.40}{1.65}
& \withvar{85.68}{1.52} & \withvar{68.51}{1.96} & \withvar{\textbf{53.10}}{2.26} & \withvar{57.90}{2.07} & 67.88 \\
& EVisRAG~\cite{sun2025visrag2}
& \withvar{67.30}{1.43} & \withvar{\underline{65.75}}{1.56} & \withvar{70.65}{1.28} & \withvar{72.92}{1.72} & \withvar{82.47}{1.16}
& \withvar{82.44}{1.08} & \withvar{\underline{72.65}}{1.47} & \withvar{45.84}{1.85} & \withvar{78.45}{1.31} & \underline{70.69} \\
\midrule
\multirow{5}{*}{\shortstack[l]{Graph/Agentic}} & RAG-Anything~\cite{guo2025raganything}
& \withvar{\underline{69.64}}{2.32} & \withvar{62.50}{2.14} & \withvar{67.57}{1.97} & \withvar{70.68}{2.48} & \withvar{\textbf{88.37}}{1.81}
& \withvar{86.89}{1.73} & \withvar{72.40}{1.92} & \withvar{39.53}{2.57} & \withvar{76.61}{2.21} & 70.60 \\
& MoLoRAG~\cite{wu2025molorag}
& \withvar{46.86}{2.66} & \withvar{65.28}{2.43} & \withvar{\underline{74.32}}{2.19} & \withvar{73.30}{2.75} & \withvar{74.88}{2.34}
& \withvar{69.17}{2.08} & \withvar{54.87}{2.49} & \withvar{44.96}{2.89} & \withvar{56.45}{2.61} & 64.32 \\
& ViDoRAG~\cite{wang2025vidorag}
& \withvar{64.03}{2.17} & \withvar{65.62}{2.02} & \withvar{66.22}{1.88} & \withvar{73.30}{2.28} & \withvar{\underline{87.79}}{1.69}
& \withvar{78.64}{1.76} & \withvar{68.18}{2.09} & \withvar{\underline{51.16}}{2.46} & \withvar{\textbf{85.48}}{1.61} & 70.05 \\
& TAP-RAG~\cite{ji2026taprag}
& \withvar{67.07}{1.92} & \withvar{64.22}{1.78} & \withvar{\textbf{74.87}}{1.54} & \withvar{\textbf{74.66}}{2.06} & \withvar{76.05}{1.68}
& \withvar{87.44}{1.46} & \withvar{64.88}{1.84} & \withvar{46.39}{2.24} & \withvar{71.75}{1.97} & 70.09 \\
& \cellcolor{black!6}\textbf{RouteDRAG}
& \cellcolor{black!6}\withvar{\textbf{74.26}}{1.12}
& \cellcolor{black!6}\withvar{\textbf{73.61}}{1.26}
& \cellcolor{black!6}\withvar{73.65}{0.98}
& \cellcolor{black!6}\withvar{73.30}{1.41}
& \cellcolor{black!6}\withvar{87.21}{0.86}
& \cellcolor{black!6}\withvar{\textbf{91.75}}{0.73}
& \cellcolor{black!6}\withvar{\textbf{77.27}}{1.05}
& \cellcolor{black!6}\withvar{48.45}{1.53}
& \cellcolor{black!6}\withvar{76.61}{1.18}
& \cellcolor{black!6}\textbf{75.86} \\
\bottomrule
\end{tabularx}
\end{table*}

\section{Experiments}
\label{sec:experiments}

We conduct extensive experiments on two multimodal document benchmarks to answer the following research questions. \textbf{Q1: Overall Performance.} How does RouteDRAG compare with existing document RAG methods in answering accuracy? \textbf{Q2: Ablation Study.} How do adaptive path routing and dynamic Agent execution contribute to performance? \textbf{Q3: Robustness Analysis.} How robust is RouteDRAG to noisy inputs, hyperparameter variations, and different model backbones? \textbf{Q4: Efficiency Analysis.} Can RouteDRAG achieve a favorable accuracy--computation trade-off compared with existing RAG methods?

\subsection{Experimental Setup}
\label{sec:setup}

\textbf{Datasets, Baselines, and Backbones.}
We evaluate RouteDRAG on two multimodal document benchmarks, DocBench~\cite{zou2025docbench} and MMLongBench-Doc~\cite{ma2024mmlongbench}. To assess its effectiveness across different retrieval and execution paradigms, we compare against four categories of baselines:
(1) direct inference using a budgeted Qwen3-VL model~\cite{bai2025qwen3vl};
(2) text-based RAG, including BM25-based sparse retrieval~\cite{robertson2009bm25} and BGE-M3-based dense retrieval~\cite{chen2024m3embedding};
(3) multimodal RAG, including M3DocRAG~\cite{cho2024m3docrag}, and EVisRAG~\cite{sun2025visrag2}; and
(4) graph-based and Agentic RAG, including RAG-Anything~\cite{guo2025raganything}, MoLoRAG~\cite{wu2025molorag}, ViDoRAG~\cite{wang2025vidorag}, and TAP-RAG~\cite{ji2026taprag}.
For a fair comparison, all baselines and RouteDRAG use the same Qwen3-VL-8B-Instruct answering model, with differences arising from their retrieval and execution strategies rather than backbone capabilities.
We additionally compare Short Path Only, Rule Routing, and Uniform Agent as controlled execution strategies. RouteDRAG employs a unified document index, frozen BGE-M3~\cite{chen2024m3embedding} for semantic representation, and Qwen3-VL-8B-Instruct~\cite{bai2025qwen3vl} for multimodal inference.

\textbf{Evaluation Metrics and Protocol.}
We report answering accuracy (Acc.), Agent usage (Agent\%), executed actions per query (Actions/Q), answering calls and tokens per query, inference latency, and peak GPU memory. Accuracy is assessed using a fixed Qwen-based judge, with failed responses counted as incorrect. For the main performance comparison, we report mean accuracy and standard deviation over five independent runs.

\begin{table*}[tbp]
\centering
\caption{Component ablation. Accuracy and Agent use are percentages; tokens are in thousands. Dashes mark the no-Agent setting. Best and second-best measured values are bold and underlined.}
\label{tab:q2_capability}
\small
\setlength{\tabcolsep}{2pt}
\renewcommand{\arraystretch}{1.16}
\begin{tabularx}{\textwidth}{@{}>{\raggedright\arraybackslash}p{0.225\textwidth}*{10}{>{\centering\arraybackslash}X}@{}}
\toprule
& \multicolumn{5}{c}{\textbf{DocBench}} & \multicolumn{5}{c}{\textbf{MMLongBench-Doc}} \\
\cmidrule(lr){2-6}\cmidrule(lr){7-11}
Method & Acc.$\uparrow$ & Agent\%$\downarrow$ & Act./Q$\downarrow$ & Calls/Q$\downarrow$ & Tok./Q$\downarrow$ & Acc.$\uparrow$ & Agent\%$\downarrow$ & Act./Q$\downarrow$ & Calls/Q$\downarrow$ & Tok./Q$\downarrow$ \\
\midrule
w/o Adaptive Path Routing
& 73.77 & \underline{15.88} & \underline{0.300} & 2.97 & 29.14
& 50.32 & \underline{16.50} & \underline{0.379} & 3.31 & 33.15 \\

w/o Dynamic Agent Execution
& 74.19 & 100.00 & 1.418 & 7.11 & 49.30
& 50.25 & 100.00 & 1.850 & 8.23 & 56.53 \\

w/o Agent Routing
& 73.41 & -- & -- & \textbf{2.01} & \textbf{24.07}
& 49.31 & -- & -- & \textbf{2.23} & \textbf{27.14} \\

All Agent Routing
& \underline{74.68} & 100.00 & 1.685 & 7.46 & 52.08
& \underline{52.52} & 100.00 & 2.148 & 8.25 & 57.02 \\

RouteDRAG-Efficient
& 74.05 & \textbf{11.52} & \textbf{0.171} & \underline{2.58} & \underline{27.36}
& 50.87 & \textbf{16.41} & \textbf{0.348} & \underline{3.20} & \underline{32.18} \\

\rowcolor{black!6}
\textbf{RouteDRAG}
& \textbf{75.86} & 78.49 & 1.349 & 6.24 & 46.04
& \textbf{53.80} & 80.02 & 1.769 & 7.02 & 50.99 \\

\bottomrule
\end{tabularx}
\par\smallskip
\end{table*}

\begin{figure}[ht]
\centering
\includegraphics[width=\columnwidth]{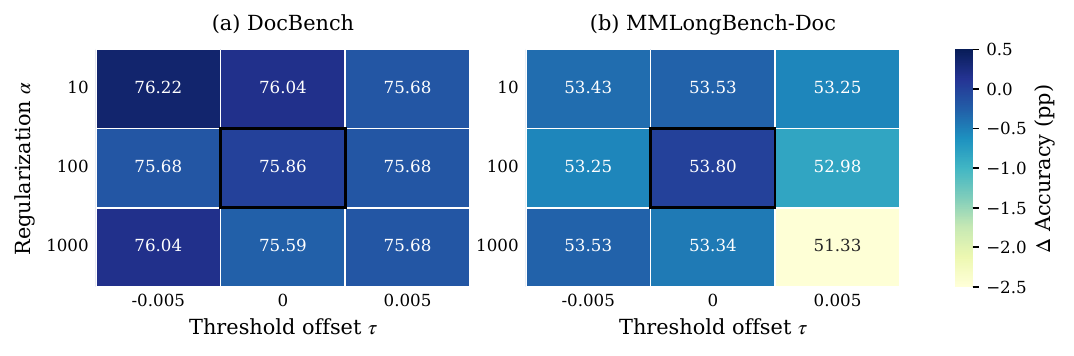}
\caption{Sensitivity to regularization strength and routing threshold on (a) DocBench and (b) MMLongBench-Doc. Cells report accuracy (\%); colors show percentage-point changes from $\alpha=100$ and zero threshold offset.}
\label{fig:alpha_tau}
\Description{Two three-by-three heatmaps show accuracy on DocBench and MMLongBench-Doc as regularization and the development-calibrated threshold offset vary.}
\end{figure}

\begin{figure*}[t]
    \centering
    \includegraphics[width=\textwidth]{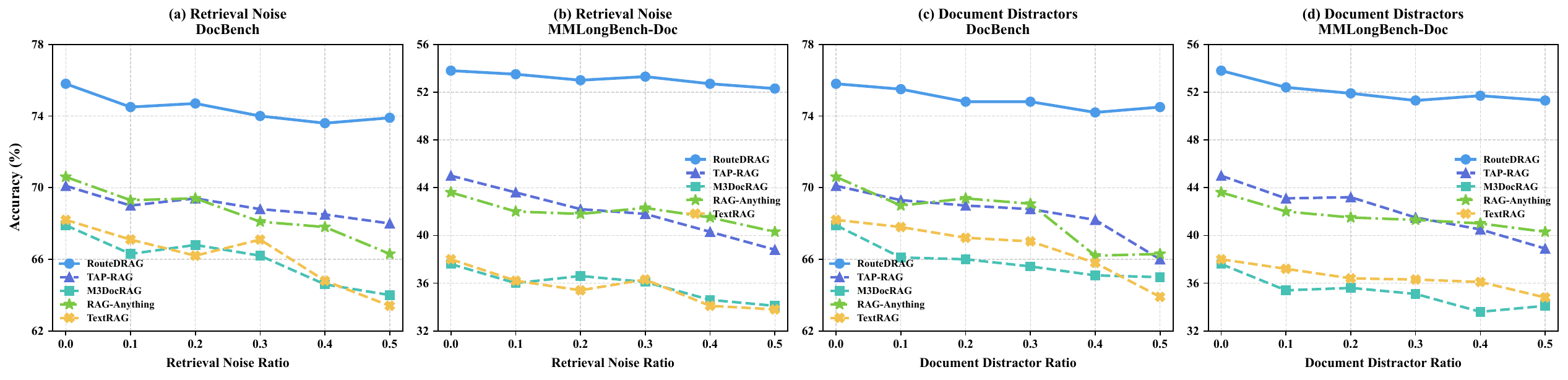}
    \caption{Robustness comparison under increasing noise ratios.
    (a--b) Retrieval noise.
    (c--d) Document distractors.}
    \label{fig:robustness}
\end{figure*}

\subsection{Overall Performance (Q1)}
\label{sec:q1}

To answer Q1, we compare RouteDRAG with representative document RAG methods across direct inference, text-based RAG, multimodal RAG, and graph-based and Agentic RAG. All experiments are conducted with five independent runs, and we report mean accuracy with standard deviations. As shown in Tables~\ref{tab:canonical_overall} and~\ref{tab:canonical_mml}, RouteDRAG achieves 75.86\% accuracy on DocBench and 53.80\% on MMLongBench-Doc, outperforming the competing RAG baselines on both benchmarks. Notably, RouteDRAG outperforms uniform Agent execution by 1.18 and 1.28 percentage points on DocBench and MMLongBench-Doc, respectively, while reducing answering calls by 16.4\% and 14.9\%. These results suggest that selective Agent invocation can
improve the accuracy--efficiency trade-off under our
evaluation settings.The gains also extend across evaluation categories: RouteDRAG achieves the best results in two of five document domains and two of four question types on DocBench, and leads across all five support-source categories and two of three support-requirement categories on MMLongBench-Doc. These results demonstrate the effectiveness of RouteDRAG across diverse document structures and question requirements, motivating further analysis of its component contributions and computational efficiency.

\begin{figure*}[t]
    \centering
    \includegraphics[width=\textwidth]{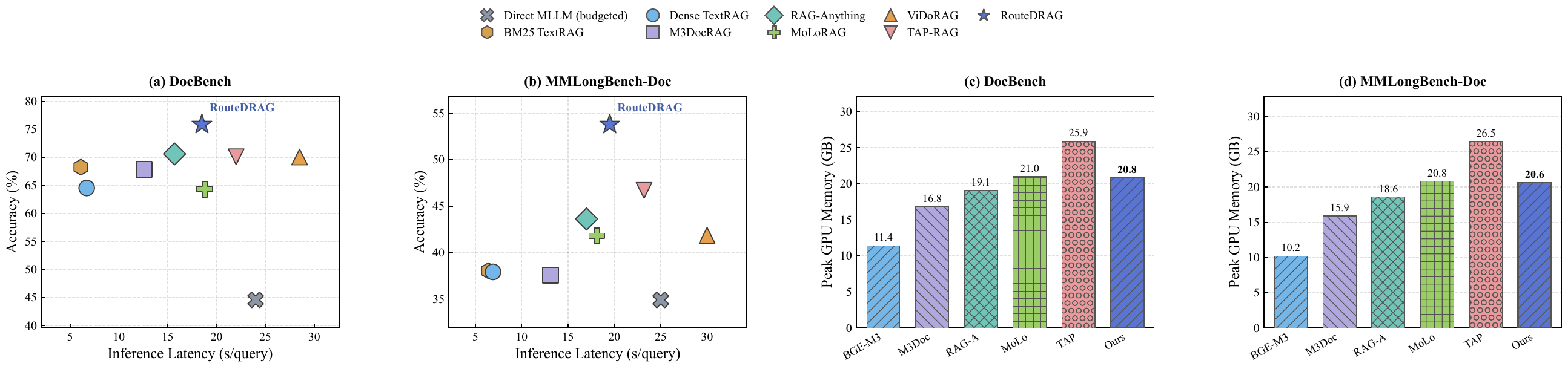}
    \caption{Practical efficiency comparison with representative RAG baselines. (a--b) Answer accuracy versus inference latency on DocBench and MMLongBench-Doc. (c--d) Peak GPU memory consumption across different methods.}
    \label{fig:efficiency_baselines}
\end{figure*}

\subsection{Ablation Study (Q2)}
\label{sec:q2}

To answer Q2, we investigate the contributions of adaptive path routing and dynamic Agent execution. As shown in Table~\ref{tab:q2_capability}, removing Agent execution reduces accuracy to 73.41\%/49.31\%, while uniform Agent execution achieves 74.68\%/52.52\%. In contrast, RouteDRAG achieves 75.86\%/53.80\% while invoking Agent for only 78.49\%/80.02\% of queries. RouteDRAG-Efficient offers a lower-cost operating point, achieving 74.05\%/50.87\% accuracy with only 11.52\%/16.41\% Agent use, exceeding rule-based routing on both datasets. These results show how selective execution balances the benefits of Agentic reasoning against unnecessary or potentially harmful operations.

\textbf{Comparison Settings.}
\emph{w/o Adaptive Path Routing} replaces learned path selection with rule-based capability assignment.
\emph{w/o Dynamic Agent Execution} assigns every query to a non-adaptive Agent execution configuration.
\emph{w/o Agent Routing} answers through the short-path portfolio, using Exact for supported structured requests and Standard or Direct VLM for the remaining queries.
\emph{All Agent Routing} sends every query to the mature Agent.
\emph{RouteDRAG} combines selective path allocation with dynamic Agent execution.
\emph{RouteDRAG-Efficient} shares the same mechanism as \emph{RouteDRAG} and yields the efficiency-oriented operating point in Table~\ref{tab:query_control} solely through a different threshold configuration.
The two all-Agent configurations compare execution policies at full Agent coverage; the routed configurations additionally expose the accuracy--cost trade-off of selective invocation.

\subsection{Robustness Analysis (Q3)}
\label{sec:q3}

To answer Q3, we examine the sensitivity of routing decisions to regularization and threshold choices, together with robustness across model backbones and noisy inputs.

\textbf{Hyperparameters.}
Fig.~\ref{fig:alpha_tau} examines the interaction between regularization strength $\alpha$ and the routing threshold. For each $\alpha$, the predictor is fitted on the training documents, and a reference threshold is calibrated on development documents. The horizontal axis varies a fixed offset from this reference; each query is then routed independently. Accuracy varies modestly across the displayed settings on DocBench. MMLongBench-Doc is more sensitive when strong regularization is combined with a positive threshold offset. Stronger regularization compresses the prediction scale, so the same offset can exclude more Agent executions. Regularization and threshold calibration should therefore be considered together.

\begin{table}[ht]
\centering
\caption{Backbone generalization on DocBench and MMLongBench-Doc. Accuracy (\%) and relative reduction in answering model calls compared with Uniform Agent (\%).}
\label{tab:backbone_current}
\resizebox{\columnwidth}{!}{%
\begin{tabular}{lcccc}
\toprule
\textbf{VLM Backbone} & \textbf{Agent} & \textbf{Random} & \textbf{Ours} & \textbf{Calls Red. (\%)} \\
\midrule
Qwen3-VL-4B
& \underline{69.60}/\underline{45.92} & 68.20/43.21
& \textbf{70.15}/\textbf{46.65} & 13.8/14.2 \\
Qwen3-VL-8B
& \underline{74.68}/\underline{52.52} & 73.66/49.88
& \textbf{75.86}/\textbf{53.80} & 16.4/14.9 \\
Qwen2.5-VL-7B
& \underline{72.78}/\textbf{48.49} & 70.40/45.92
& \textbf{73.32}/\underline{48.30} & 14.1/14.4 \\
InternVL3-8B
& \underline{73.50}/\underline{50.23} & 71.10/47.59
& \textbf{74.23}/\textbf{51.05} & 15.0/14.3 \\
Gemma-3-12B
& \textbf{71.96}/\underline{47.57} & 69.54/45.97
& \underline{71.87}/\textbf{48.49} & 13.7/15.2 \\
\bottomrule
\end{tabular}%
}
\end{table}

\textbf{Backbone Generalization.}
Table~\ref{tab:backbone_current} evaluates RouteDRAG across multimodal backbones of different scales and architectures. With Qwen3-VL-4B, RouteDRAG achieves 70.15\%/46.65\% accuracy on DocBench/MMLongBench-Doc, improving over uniform Agent execution by 0.55/0.73 percentage points while reducing answering calls by 13.8\%/14.2\%. With InternVL3-8B, it improves accuracy by 0.73/0.82 percentage points. Gains vary across backbones.

\textbf{Robustness to Noisy Information.}
We consider two perturbations: \emph{Retrieval Noise}, which adds irrelevant content to retrieved results, and \emph{Document Distractors}, which adds unrelated pages to documents. Fig.~\ref{fig:robustness} reports accuracy as the perturbation ratio increases from 0 to 0.5. At the highest ratio, RouteDRAG achieves 73.9\%/52.3\% accuracy on DocBench/MMLongBench-Doc under Retrieval Noise and 74.5\%/51.3\% under Document Distractors. It outperforms TAP-RAG by 5.9/13.5 and 8.5/12.4 percentage points under the two perturbations, respectively.

\subsection{Efficiency Analysis (Q4)}
\label{sec:q4}

To answer Q4, we evaluate answering calls and tokens, inference latency, and peak GPU memory consumption.

\textbf{Efficiency of Selective Execution.}
Table~\ref{tab:q2_capability} shows that RouteDRAG uses 6.24/7.02 answering calls and 46.04k/50.99k tokens per query on DocBench/MMLongBench-Doc. Compared with \textit{All Agent Routing}, it improves accuracy by 1.18/1.28 percentage points while reducing calls by 16.4\%/14.9\% and tokens by 11.6\%/10.6\%. Compared with \textit{w/o Dynamic Agent Execution}, it improves accuracy by 1.67/3.55 percentage points while reducing both calls and tokens. The lower costs of \textit{w/o Adaptive Path Routing} and \textit{w/o Agent Routing} come with reduced accuracy.

\textbf{Practical Efficiency Comparison.}
Fig.~\ref{fig:efficiency_baselines}(a--b) compares answer accuracy against inference latency on DocBench and MMLongBench-Doc; panels (c--d) compare peak GPU memory consumption. These measurements complement the answering-call and token costs in Table~\ref{tab:q2_capability}.
\section{Conclusion}
\label{sec:conclusion}

In this paper, we investigated a fundamental question in Agentic document RAG: when is Agentic execution necessary, and how should its operations be controlled? Our empirical findings reveal that Agentic execution provides query-dependent benefits, while additional operations yield gains, ties, and degradations in answer correctness. Motivated by these observations, we proposed \textbf{RouteDRAG}, a framework that combines adaptive path routing with dynamic Agent execution through sequential policy learning. By learning from observed execution outcomes, RouteDRAG selectively invokes Agent and refines its subsequent operations. Experiments on two multimodal document benchmarks demonstrate improved accuracy with fewer answering calls compared with uniform Agent execution. These results highlight the importance of controlling not only whether an Agent is invoked, but also how it executes. While this work focuses on single-turn document question answering, extending selective execution to multi-turn document interactions presents a promising direction for future research.

\FloatBarrier
\clearpage
\nobalance
\bibliographystyle{ACM-Reference-Format}
\clearpage
\makeatletter
\global\@colroom=\@colht
\global\vsize=\@colht
\makeatother
\bibliography{references}
\clearpage
\appendix
\setcounter{table}{0}
\renewcommand{\thetable}{A\arabic{table}}
\setcounter{figure}{0}
\renewcommand{\thefigure}{A\arabic{figure}}
\setcounter{algorithm}{0}
\renewcommand{\thealgorithm}{A\arabic{algorithm}}
\section{Experimental Details}
\label{app:judge}
\subsection{Datasets, Readers, and Execution Paths}
\label{app:allocation_protocol}

We evaluate RouteDRAG on all 1,102 DocBench questions from 229 documents and 1,091 MMLongBench-Doc questions from 135 documents. All paths share the document index and frozen BGE-M3 representations, using Qwen3-VL-8B-Instruct with an FP8 backend; final Agent synthesis is capped at 8,192 output tokens. The designated short path uses Standard or eligible Direct VLM, while the \emph{Actor} denotes the Agent with its native Planner, Workspace, legal-action filtering, and learned intervention. For controlled allocation comparisons, we reuse the recorded short-path and Actor answers for each query, keeping Exact Specialist outputs fixed across policies.

\subsection{Baseline Configurations}
RAG-Anything uses upstream mixed retrieval with entity top-$k=50$ and chunk top-$k=10$. It reuses the canonical graph index and uses a Qwen FP8 service with a one-sentence answer instruction. MoLoRAG retains native multimodal retrieval and traversal. BM25 and BGE-M3 share native MinerU text chunks, including native table text, with document-local top-20 retrieval and a one-sentence Qwen FP8 reader. The BGE-M3 baseline retrieves native text chunks using frozen embeddings. 

M3DocRAG follows the comparison adapter: document-local ColPali \cite{faysse2025colpali} retrieval selects three pages, rendered at 144 DPI for the Qwen FP8 reader.

Budgeted Direct MLLM reads all pages up to 128. For the remaining baselines, we preserve their original retrieval and execution procedures while using the same Qwen3-VL-8B-Instruct answering model and evaluation protocol across all methods.

\subsection{Robustness Evaluation Protocol}
\label{app:noise_robustness}

We evaluate RouteDRAG and representative RAG baselines on DocBench and MMLongBench-Doc under two types of perturbations, with noise ratios $\rho \in \{0,0.1,0.2,0.3,0.4,0.5\}$. A ratio of zero denotes the original setting. Distractors are sampled from other documents in the same benchmark, excluding the source document of the current query. The questions, reference answers, model parameters, and routing thresholds remain fixed throughout evaluation.

\textbf{Retrieval Noise.}
For each query, we obtain the native retrieval results of each method and denote the number of retrieved units by $K$. We then append $\lfloor \rho K \rfloor$ irrelevant units sampled from the distractor pool, preserving the original results. Thus, $\rho=0.5$ adds distractors equal to half the original number of retrieved units, while preserving the original retrieved content. The document index and retrieval model remain unchanged, isolating interference introduced at retrieval time.

\textbf{Document Distractors.}
For a document containing $P$ pages, we append $\lfloor \rho P \rfloor$ unrelated pages sampled from other documents, inserting them at randomly selected positions while preserving the relative order of the original pages. At $\rho=0.5$, the number of added pages is half the original page count. Each method processes the perturbed documents using its native indexing pipeline, including index reconstruction where required. This setting evaluates the effect of irrelevant pages present before retrieval.

\subsection{Inference Efficiency Measurement}
\label{app:efficiency_protocol}

We measure inference latency and peak GPU memory for RouteDRAG
and representative RAG baselines under the same hardware and
runtime conditions. All methods use batch size 1, with five
warm-up queries followed by 100 timed queries drawn from the
same question set. We report the mean end-to-end latency per
query and peak GPU memory usage. Experiments run on a single
NVIDIA A100 GPU, with inference precision and backend following
each method's configuration.

End-to-end latency includes online retrieval, routing, Agent
planning and tool execution, and final answer generation.
Offline document parsing and index construction are excluded.
Peak GPU memory includes model weights and runtime allocations.
All methods retain their designated retrieval and answering
pipelines; RouteDRAG additionally includes routing and learned
execution-control overhead.

\subsection{Routing Parameters}
\label{app:routing_parameters}

For each query, a fixed policy first designates Standard or
Direct VLM as the short path $C(q)$. The learned router then
predicts the signed benefit of Agent execution relative to
$C(q)$, using the query representation and short-path identity.
Agent is invoked when the predicted benefit exceeds the fixed
threshold $\tau=0.02$; otherwise, $C(q)$ is executed.

The router is trained on paired Agent and short-path outcomes.
Its weights are fitted on training data, while model configurations
and routing thresholds are selected using development data.
The resulting routing function is applied independently to each
query with its fixed threshold.

\subsection{Document-Disjoint Cross-Validation}
\label{app:router_cross_validation}

We evaluate the learned Controller and router using five-fold
cross-validation at the source-PDF level, stratified by dataset.
Questions from the same PDF are assigned to the same outer fold,
with duplicate copies grouped under a common PDF identity. In
each round, one fold serves as the test set. Among the remaining
documents, 20\% form an inner development set and the rest form
the training set. The resulting training/development/test
proportions are approximately 64\%/16\%/20\% of documents.
Question counts vary with the number of questions associated
with each document.

In each round, the shared projection and operation Controller
are trained using operation-level outcomes from the training
partition, with checkpoint selection based on the inner
development set. They are then frozen, and a separate router
is fitted using the training questions and their paired
correctness targets,
$\Delta_i=Y(A_\phi(q_i))-Y(C(q_i))$.
The inner development set determines the router configuration
and decision threshold. The trained components and selected
threshold remain fixed when applied to the outer test fold.

We pool the predictions from the five outer test folds, so each
query in the routing evaluation contributes one out-of-fold
routing decision. Dataset-level routing results are aggregated
from these held-out predictions.

\section{Method Implementation Details}
\label{app:method_details}

\subsection{Shared Index and Source References}
\label{app:index_details}

Figure~\ref{fig:routedrag_framework} distinguishes the vector index
$I_D$, the entity--relation graph $\mathcal{G}_D$, and source
references $\rho_D$. The vector index supports similarity retrieval
of text chunks and entity descriptions, while graph relations enable
connected retrieval. Visual descriptions are indexed with links to
their original pages and image assets. Source references retain
document-local identifiers and locations, allowing retrieved
descriptions to be traced to their source tables or figures.
Retrieving a visual description and inspecting its original image
are separate operations.

Frozen semantic vectors have 1,024 dimensions. A shared learned
projection maps query, chunk, and entity representations to 64
dimensions. Each input contains up to eight chunk vectors and four
entity vectors, with unavailable entries padded and masked before
attention. The graph branch pools retrieved entity representations,
while relations remain available for graph retrieval. Lightweight
statistics describe query form, available content, and query--content
similarity; these features are computed from information available
at each decision point.

\subsection{Exact and Short-Path Interfaces}
\label{app:short_path_details}

Exact is a deterministic metadata gate that returns an answer only
when a supported structured request satisfies the required
completeness checks. Otherwise, the fixed short-path policy selects
Standard by default or Direct VLM when visual inspection is eligible
for the query. Standard performs graph-backed retrieval and answering.
Direct VLM first applies a text precheck, invokes visual reading when
needed, and falls back to text when visual reading is unnecessary.

The designated short path is denoted by $C(q)$. Adaptive Path Routing predicts whether the Agent offers sufficient benefit over the designated shorter path $C(q)$ to justify additional execution. Exact outputs remain fixed across allocation
policies, and the Agent, denoted as the \emph{Actor}, retains its
native Planner, Workspace, legal-action constraints, and tool
interfaces.

\subsection{Workspace, Legal Actions, and Tool Arguments}
\label{app:execution_details}

The query-specific Workspace stores retrieved text, entities,
relations, source references, tool outcomes, and action history.
Each tool result is committed with its provenance, and duplicate
content is removed. Initial retrieval creates the starting state;
subsequent operations update the same Workspace while reusing the shared document index.

At step $t$, execution rules construct the legal action set
$\Gamma_t$ from the available resources and current Workspace.
Candidate operations include text search, graph expansion, structured
lookup, visual inspection, and \texttt{FINISH} when permitted.
The learned Controller scores only legal actions, while the Planner
proposes an action and its arguments. If the proposed action is
retained, its arguments are reused; otherwise, the Planner generates
arguments for the selected tool.

The Controller receives projected query and content representations,
state statistics, operation embeddings, and an eight-dimensional
history vector comprising the step, history length, five operation
counts, and failure count, transformed by $\log(1+x)$. Operation
embeddings have 64 dimensions. The action-value head uses a
32-unit hidden layer and a scalar output, with \texttt{FINISH} as
the reference and a fixed refinement margin of $m=0.15$.
The relative scores estimate operation-level correctness gains against the FINISH reference. After a permitted
stop or execution limit, the original synthesis procedure generates
the final answer from the accumulated content and source references.

\subsection{Learning Signals}
\label{app:learning_details}

The labels $Y_{\mathrm{stop}}$ and $Y_{\mathrm{continue}}$ in
Figure~\ref{fig:routedrag_framework} denote answer correctness before
and after one additional operation, respectively, as defined in
Equation~\ref{eq:operation_gain}. Only evaluated state--action pairs
contribute to the regression and ranking losses. Planner imitation
provides an auxiliary signal that preserves the original Planner's
operation preferences.

At the path level, $r_A=Y(A_\phi(q))$ and $r_C=Y(C(q))$ denote the
paired correctness outcomes of the Actor and designated short path.
Their difference, $r_A-r_C$, provides the signed routing target in
Equation~\ref{eq:paired_route_target}. The shared projection and
Controller are trained first and then frozen while the routing heads
learn from the updated Actor's outcomes. Reference answers and
correctness labels provide offline supervision; inference uses only
the learned representations, predicted scores, and legal execution
rules.

\begin{figure*}[t]
\centering

\begin{minipage}[t]{0.64\textwidth}
    \vspace{0pt}
    \centering
    \includegraphics[width=\linewidth]
        {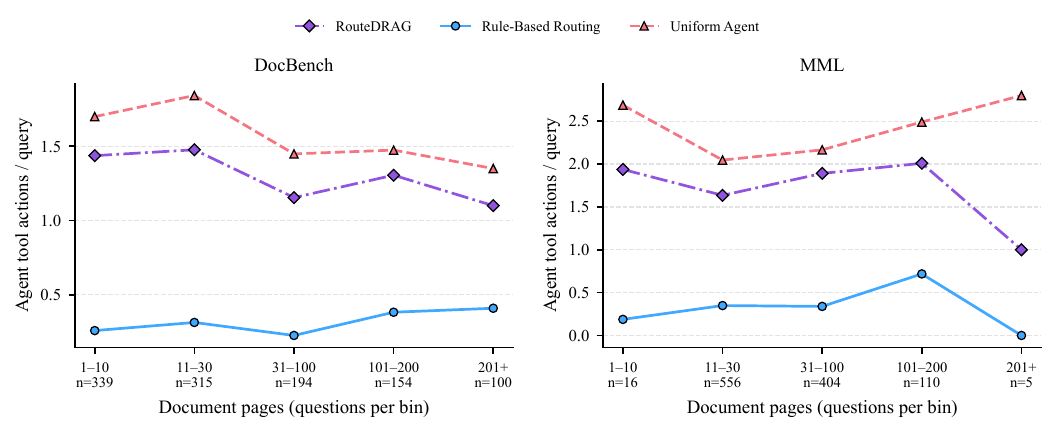}
    \caption{Tool use across document-length groups. Counts show
    questions per bin; the MML 201+ bin contains five questions from
    one document.}
    \label{fig:scale}
    \Description{Two dataset panels compare RouteDRAG, rule routing,
    and uniform Agent actions across page-count groups.}
\end{minipage}
\hfill
\begin{minipage}[t]{0.34\textwidth}
    \vspace{0pt}
    \centering
    \includegraphics[width=\linewidth]
        {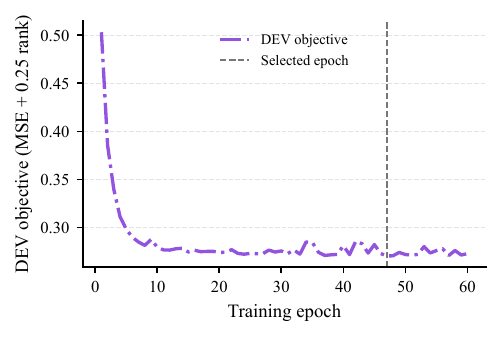}
    \caption{Benefit-predictor validation loss. The marker identifies
    the selected epoch, 47.}
    \label{fig:router_convergence}
    \Description{Validation loss across training epochs with the
    selected epoch marked.}
\end{minipage}

\end{figure*}

\subsection{Training and Policy Selection}
\label{app:training_selection}

\textbf{Sequential Training.}
The document index, semantic encoder, and language-model
backbones remain frozen. In the first stage, the shared
projection and operation Controller are trained using observed
operation-level outcomes through regression, same-state ranking,
and auxiliary Planner imitation. In the second stage, the shared
projection and Controller are frozen, and the routing heads
are trained using the paired correctness outcomes of the Actor
and the designated short path. This two-stage procedure is
repeated independently in each fold, using the training,
development, and test partitions defined by source PDF in
Appendix~\ref{app:router_cross_validation}.

\textbf{Checkpoint and Threshold Selection.}
Each stage selects its checkpoint using the corresponding
development objective. The operation-refinement margin is fixed
at $m=0.15$. The per-query routing threshold $\tau$ is selected
on the development set, and both model parameters and
thresholds remain fixed throughout test evaluation.

\textbf{Execution Policies.}
In per-query inference, the router invokes the Actor when the
predicted relative benefit $\widehat{\Delta}_\theta(q)$ exceeds
$\tau$; otherwise, it uses the designated short path. Once
invoked, the Controller refines the Planner's proposal only
when a legal alternative exceeds the proposed action's score
by the fixed refinement margin.

The high-accuracy configuration uses the shared router and
fixed threshold specified in Appendix~\ref{app:routing_parameters}.
Each query is routed independently: the Agent is invoked when
$\widehat{\Delta}(q)>\tau$; otherwise, the designated short path
is used.

\section{Supplementary Execution Profiles}

\subsection{Natural Document-Length Profile}
\label{app:length}

Figure~\ref{fig:scale} groups the evaluated documents by page count and
reports tool actions per query. RouteDRAG uses fewer actions than uniform
execution across the displayed groups. Tool use also varies within the
length range, reflecting query requirements as well as document size.
Question counts accompany each bin; the longest MML bin contains five
questions from one PDF.

\subsection{Validation-Loss Trace}
\label{app:router_convergence}

Figure~\ref{fig:router_convergence} plots the benefit predictor's
validation objective: signed-benefit regression plus a ranking term
weighted by 0.25. Training uses paired execution outcomes, and validation
selects epoch 47. The trace describes optimization of the routing
objective; the workload and document-disjoint allocation results are
reported separately.

\subsection{Per-Query Routing Operating Points}
\label{app:routing_operating_points}

Table~\ref{tab:query_control} compares four routing configurations
with fixed thresholds calibrated on the development set.
RouteDRAG-Efficient, RouteDRAG-Medium, RouteDRAG-High, and RouteDRAG share the
accuracy-oriented router with different thresholds.
RouteDRAG-Efficient achieves accuracies of 74.05\% and 50.87\%
on DocBench and MMLongBench-Doc, respectively, while limiting
Agent usage to 11.52\% and 16.41\%, with 2.58 and 3.20 calls per
query. Increasing the routing intensity improves accuracy at
the cost of higher resource usage. RouteDRAG-Medium achieves
74.73\%/52.29\% accuracy at 42.65\%/57.56\% Agent usage,
while RouteDRAG-High reaches 75.18\%/52.93\% accuracy at
59.53\%/71.13\% Agent usage on the two benchmarks, respectively.
RouteDRAG achieves the highest accuracy, at 75.86\% on DocBench
and 53.80\% on MMLongBench-Doc, but also incurs the highest
Agent usage (78.49\%/80.02\%) and the most calls per query
(6.24/7.02). Overall, these operating points illustrate the
trade-off between accuracy and inference cost enabled by
routing decisions.

\begin{table}[t]
\centering
\caption{Accuracy--cost trade-offs of per-query routing configurations. Accuracy and actual Agent usage are percentages.}
\label{tab:query_control}
\small
\setlength{\tabcolsep}{3pt}
\resizebox{\columnwidth}{!}{%
\begin{tabular}{@{}lrrrrrr@{}}
\toprule
& \multicolumn{3}{c}{DocBench}
& \multicolumn{3}{c}{MMLongBench-Doc} \\
\cmidrule(lr){2-4}\cmidrule(lr){5-7}
Configuration
& Acc.$\uparrow$ & Agent\%$\downarrow$ & Calls/Q$\downarrow$
& Acc.$\uparrow$ & Agent\%$\downarrow$ & Calls/Q$\downarrow$ \\
\midrule
RouteDRAG-Efficient
& 74.05 & 11.52 & 2.58
& 50.87 & 16.41 & 3.20 \\
RouteDRAG-Medium
& 74.73 & 42.65 & 4.22
& 52.29 & 57.56 & 5.72 \\
RouteDRAG-High
& 75.18 & 59.53 & 5.09
& 52.93 & 71.13 & 6.57 \\
RouteDRAG
& 75.86 & 78.49 & 6.24
& 53.80 & 80.02 & 7.02 \\
\bottomrule
\end{tabular}%
}
\end{table}

\clearpage
\end{document}